# Outliers resistant image classification by anomaly detection

Anton Sergeev
*National Research University*
*Higher School of Economics*
Moscow, Russia
avsergeev@hse.ru

Victor Minchenkov
*National Research University*
*Higher School of Economics*
Moscow, Russia
vminchenkov@hse.ru

Aleksei Soldatov
*National Research University*
*Higher School of Economics*
Moscow, Russia
avsoldatov@edu.hse.ru

Vasiliy Kakurin
*National Research University*
*Higher School of Economics*
Moscow, Russia
vvkakurin@mail.ru

Yaroslav Mazikov
*National Research University*
*Higher School of Economics*
Moscow, Russia
yaamazikov@edu.hse.ru

## Abstract

Various technologies, including computer vision models, are employed for the automatic monitoring of manual assembly processes in production. These models detect and classify events such as the presence of components in an assembly area or the connection of components. A major challenge with detection and classification algorithms is their susceptibility to variations in environmental conditions and unpredictable behavior when processing objects that are not included in the training dataset. As it is impractical to add all possible subjects in the training sample, an alternative solution is necessary. This study proposes a model that simultaneously performs classification and anomaly detection, employing metric learning to generate vector representations of images in a multidimensional space, followed by classification using cross-entropy. For experimentation, a dataset of over 327,000 images was prepared. Experiments were conducted with various computer vision model architectures, and the outcomes of each approach were compared.

## Аннотация

Различные технологии, включая модели компьютерного зрения, применяются для автоматизированного контроля процессов ручной сборки на производстве. Эти модели позволяют обнаруживать и классифицировать события, такие как наличие компонентов в области сборки или их соединение. Основной проблемой алгоритмов детекции и классификации является их чувствительность к изменениям условий окружающей среды и непредсказуемое поведение при обработке объектов, отсутствующих в обучающей выборке. Поскольку включение всех возможных объектов в обучающую выборку является непрактичным, требуется альтернативное решение. В данном исследовании предлагается модель, одновременно выполняющая задачи классификации и детекции аномалий. Модель использует метод metric learning для построения векторных представлений изображений в многомерном пространстве с последующей классификацией с помощью функции перекрестной энтропии. Для проведения экспериментов был подготовлен набор данных, включающий более 327 000 изображений. Эксперименты проводились с различными архитектурами моделей компьютерного зрения, и результаты каждого подхода были сравнены.

1. Введение

В процессе ручной сборки на сборочном столе одновременно находятся многочисленные объекты, включая детали, инструменты и частично собранные механизмы. Для автоматического контроля процесса требуется решать задачи детекции и классификации различных событий. В частности, требуется определять все имеющиеся на сборочном столе детали, инструменты, частично собранные механизмы, и отличать правильно и неправильно выполненные этапы сборки. Например, при обычном соединении некоторого винта с остальным механизмом, требуется проверить, правильной ли стороной был присоединен винт. Такие проверки помогают избежать ошибок ручной сборки, возникающих вследствие человеческого фактора – невнимательности, усталости или неопытности работника. Также, помимо контроля правильности сборки можно дополнительно следить за выполнением техники безопасности.

2. Постановка задачи

Наша система контролирует процесс ручной сборки некоторого механизма. Над сборочным столом расположена камера, передающая видео на вход нейросети. На сборочном столе находятся различные объекты, среди которых могут быть детали для сборки механизма, уже частично собранный механизм, инструменты, и, возможно, посторонние объекты (например, детали от другого механизма, перчатки, и т.д.). Мы обучаем нейросеть детектировать все объекты на столе, определять, относятся ли они к сборочному процессу, а также определить их классы.

Для решения поставленной задачи мы предлагаем двухэтапный подход: модель детекции отвечает за обнаружение объектов (в том числе посторонних) на сборочном столе, и затем объекты классифицируются.

При использовании обычных моделей детекции и классификации изображений (например, YOLOv5 [1]), возникают проблемы с классификацией объектов, не присутствовавших в обучающей выборке. Нередко такие модели с высокой уверенностью относят посторонний объект к одному из классов, что может вызвать ложное срабатывание нашей системы, и, как следствие, пропуск нарушения сборочного процесса. Так как на практике невозможно собрать в тренировочном множестве все возможные объекты, которые могут в будущем оказаться на сборочном столе, мы решили модифицировать постановку задачи классификации.

Чтобы обучить устойчивую к различным шумам и посторонним объектам модель классификации, мы предлагаем на этапе обучения одновременно решать две задачи: детекции аномалий и классификации.

## 3. Обзор литературы

Методы классификации объектов, позволяющие отличать объекты разных классов, имеющих схожие признаковые представления, находит применение в различных областях, включая распознавание лиц, обнаружение вредоносных программ, анализ медицинских данных, выявление патологий и обнаружение мошенничества. С развитием технологий машинного обучения их применение в таких задачах стало повсеместным, позволяя существенно повысить качество работы систем. В подобных задачах активно используются различные методы машинного обучения, решающие задачи классификации, кластеризации и обнаружения аномалий, а также метод metric learning.

Мы решили использовать метод metric learning, сочетающий в себе сильные стороны других методов, и не требующий хранения в памяти больших наборов данных при работе системы, только самой нейросетевой модели. Благодаря этому, вычислительная сложность обработки одного кадра не зависит от конкретной задачи, только от архитектуры используемой модели и оборудования, что позволяет более предсказуемо оценивать расходы заказчика.

### 3.1. Подход Deep Metric Learning

Подход Deep Metric Learning заключается в решении задачи metric learning с помощью глубоких нейронных сетей. Модель строит векторные представления входных изображений, которые затем применяются для решения поставленной нами задачи. Для применения этого подхода не требуется дополнительной разметки данных (помимо меток классов), что значительно облегчает его применение для обучения моделей по сравнению с другими методами, такими как predefined stripe segmentation [14], multi-scale fusion [15], global-local feature learning [16].

Наиболее часто используемые функции потерь: classification loss [7] [8], verification loss [9] [10], contrastive loss [5] [11], triplet loss [6] [12] и quadruplet loss [2] [13].

Мы применяем quadruplet loss, обладающую на практике большей обобщающей способностью, чем другие функции потерь, согласно работам Weihua и др. [2], Yan и др. [13].

Данная функция потерь минимизирует внутриклассовые расстояния и максимизирует межклассовые, рассматривая четверки объектов $(a, p, n_1, n_2)$. Здесь $a$ – опорный объект класса $y_i$, $p$ – другой объект класса $y_i$, $n_1$ - объект класса $y_j, j \neq i$, $n_2$ - объект класса $y_k, k \neq i, k \neq j$, и функция потерь следующая:

$$L_{quad} = \sum_{a,p,n_1} [\rho(a,p) - \rho(a,n_1) + m_1] + \sum_{a,p,n_1,n_2} [\rho(a,p) - \rho(n_1,n_2) + m_2]$$

где $m_1, m_2$ – неотрицательные пороги (гиперпараметр).

## 4. Данные

Для обучения наших моделей мы собрали даттасеты с изображениями деталей, инструментов и этапов сборок различных объектов, а также различные изображения посторонних объеков (например, фотографии животных). При обучении моделей на каждом из наших датасетов, качестве изображений посторонних объектов мы рассматривали изображения из всех остальных датасетов.

Всего, мы получили более 327K изображений в тренировочном множестве, 48K в валидационном и 29K в тестовом. Примеры наших датасетов на Рис. 1, 2.

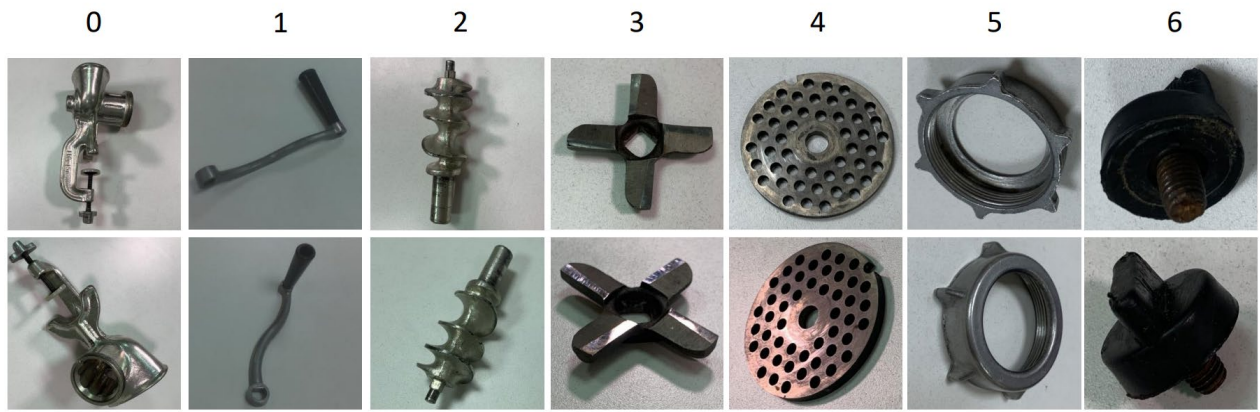

Рис. 1. Набор данных "Grinder Details"

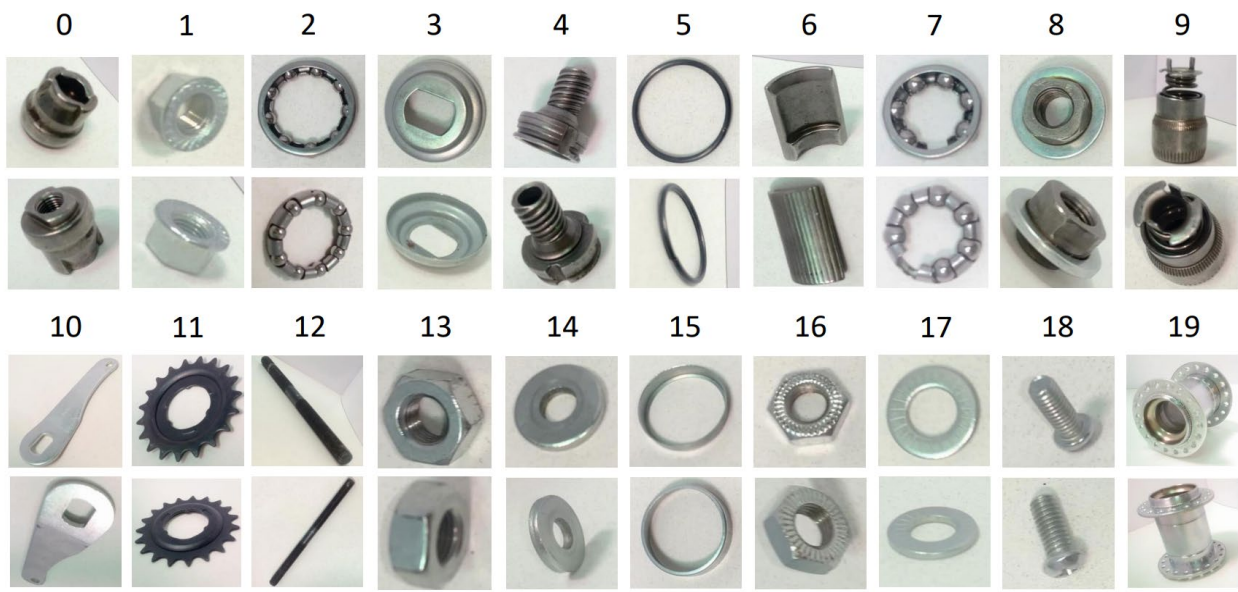

Рис. 2. Набор данных "Bicycle Parts"

## 5. Описание модели

Для построения векторных представлений изображений мы использовали различные предобученные модели компьютерного зрения, а для решения задачи последующей классификации векторов мы применили к ним небольшую полносвязную сеть.

Формально, рассмотрим набор изображений $X \subset R^D$, для наших моделей $D = 3 \times 192 \times 192$. Помимо изображений, рассмотрим их разметку на классы $Y$, где $Y \subset \{-1, 0, \ldots C-1\}$. Класс "-1" – специальный класс для изображений, не относящихся к решаемой задаче (посторонних объектов).

Наша модель в процессе обучения учит отображение $X \mapsto \hat{X} \subset R^d, d \ll D$, понижающее размерность признакового пространства. Помня о проблеме «проклятия размерности» в многомерных векторных пространствах, мы не можем задавать большие значения d. Также, стоит отметить, что в случае решения задачи классификации векторных представлений с помощью полносвязной сети, необходимо соблюдать условие $d \leq C + 1$. Иначе, число классов будет меньше размерности вектора, что может привести к плохо обусловленным весовым матрицам и сильному переобучению сети. В наших моделях, в зависимости от числа классов в датасете, мы меняли значение $d$ от 8 до 32.

Построенные таким образом векторные представления объектов мы применяем как для решения задачи детекции аномалий, так и для решения задачи классификации.

Используемая нами функция потерь Quadruplet Loss минимизирует внутриклассовые расстояния и максимизирует межклассовые, по сути, решая задачу детекции аномалий на основе попарных расстояний. Так как в нашей задаче, кроме $C$ классов из разметки имеется дополнительный класс посторонних изображений, мы модифицировали стандартную реализацию Quadruplet Loss. В частности, в качестве опорного (a) и позитивного (p) примеров может быть выбран любой объект одного из C классов (не включая посторонние объекты) в качестве негативного ($n_1$) примера – объект любого из C+1 классов (включая посторонние объекты), и в качестве четвертого ($n_2$) примера - объект любого из C+1 классов, исключая классы объектов $(a, p, n_1)$.

Для решения задачи классификации векторных представлений мы применяем полносвязную нейронную сеть, реализующую отображение $\hat{X} \mapsto R^{C+1}$, и с помощью функции SoftMax преобразуем полученные C+1 чисел в вероятности принадлежности к соответствующим классам.

## 5.1. Архитектура модели

Мы рассматривали несколько предобученных моделей[1] компьютерного зрения: Swin Transformer, MobileNetV3, EfficientNetV2. Более подробная информация о соответствующих моделях отражена в таблице 1.

| Предобученные параметры | Accuracy@1 | Accuracy@5 | Params | GFLOPS |
|---|---|---|---|---|
| Swin_T_Weights.IMAGENET1K_V1 | 81.474 | 95.776 | 28.3M | 4.49 |
| MobileNet_V3_Large_Weights.IMAGENET1K_V2 | 75.274 | 92.566 | 5.5M | 0.22 |
| EfficientNet_V2_S_Weights.IMAGENET1K_V1 | 84.228 | 96.878 | 21.5M | 8.37 |

Таблица 1. Предобученные модели

Для решения задачи классификации векторных представлений, имеющих размерность $d$, мы применили небольшую полносвязную сетью. Так как используемая предобученная модель компьютерного зрения строит векторные представления, пригодные для решения задач кластеризации и детекции аномалий (исходя из оптимизируемой функции потерь Quadruplet Loss), мы решили отказаться от сложных архитектур для слоя классификации. Таким образом, при оптимизации функции потерь классификации мы учим предобученную модель оптимизировать Quadruplet Loss, дополнительно требуя пригодности полученных векторов для классификации простой нейронной сетью.

В итоге, мы использовали двухслойную полносвязную сеть с батч-нормализацией и функцией активации LeakyReLU, имеющую внутреннюю размерность $d$.

## 5.2. Функции потерь

---
[1] https://pytorch.org/vision/stable/models.html#table-of-all-available-classification-weights

Для решения задачи детекции аномалий на основе попарных расстояний между векторными представлениями изображений, мы взяли функцию потерь QuadrupletLoss [2]. Далее, эта функция потерь обозначается $L_{quad}$. Дополнительно, мы применили метод hard examples mining. Он позволяет ввести дополнительный штраф для наиболее сложных для модели пар объектов.

Для решения задачи классификации мы применяли функцию потерь FocalLoss [3] - модификацию CrossEntropyLoss, с гиперпараметром, позволяющим ввести дополнительный штраф на наиболее сложных для модели объектах. Эта функция потерь обозначается $L_c$.

В качестве итоговой функции потерь применялась $L = L_{quad} + w_c \cdot L_c$ – взвешенная сумма двух функций потерь.

Так как в наших датасетах различное число классов – от 6 до 47, мы использовали коэффициент $w_c$ для нормировки функции потерь классификации. Если модель предсказывает равные вероятности для всех $C$ классов, значение функции потерь для одного объекта будет равно $\ln(C)$, поэтому мы взяли нормировочный коэффициент $w_c = \frac{1}{\ln(C)}$.

## 6. Эксперименты

### 6.1. Подготовка данных

Для увеличения размеров тренировочного корпуса изображений мы применяли следующие аугментации: афинные преобразования, зеркальное отражение, изменение яркости и контраста, MotionBlur, ISONoise, уменьшение разрешения изображения, RandomAutocontrast, RandomEqualize, RandomPosterize, ColorJitter, гауссовский шум и изменение гистограммы освещённости. Примеры аугментированных изображений из тренировочных множеств изображены на Рис. 3.

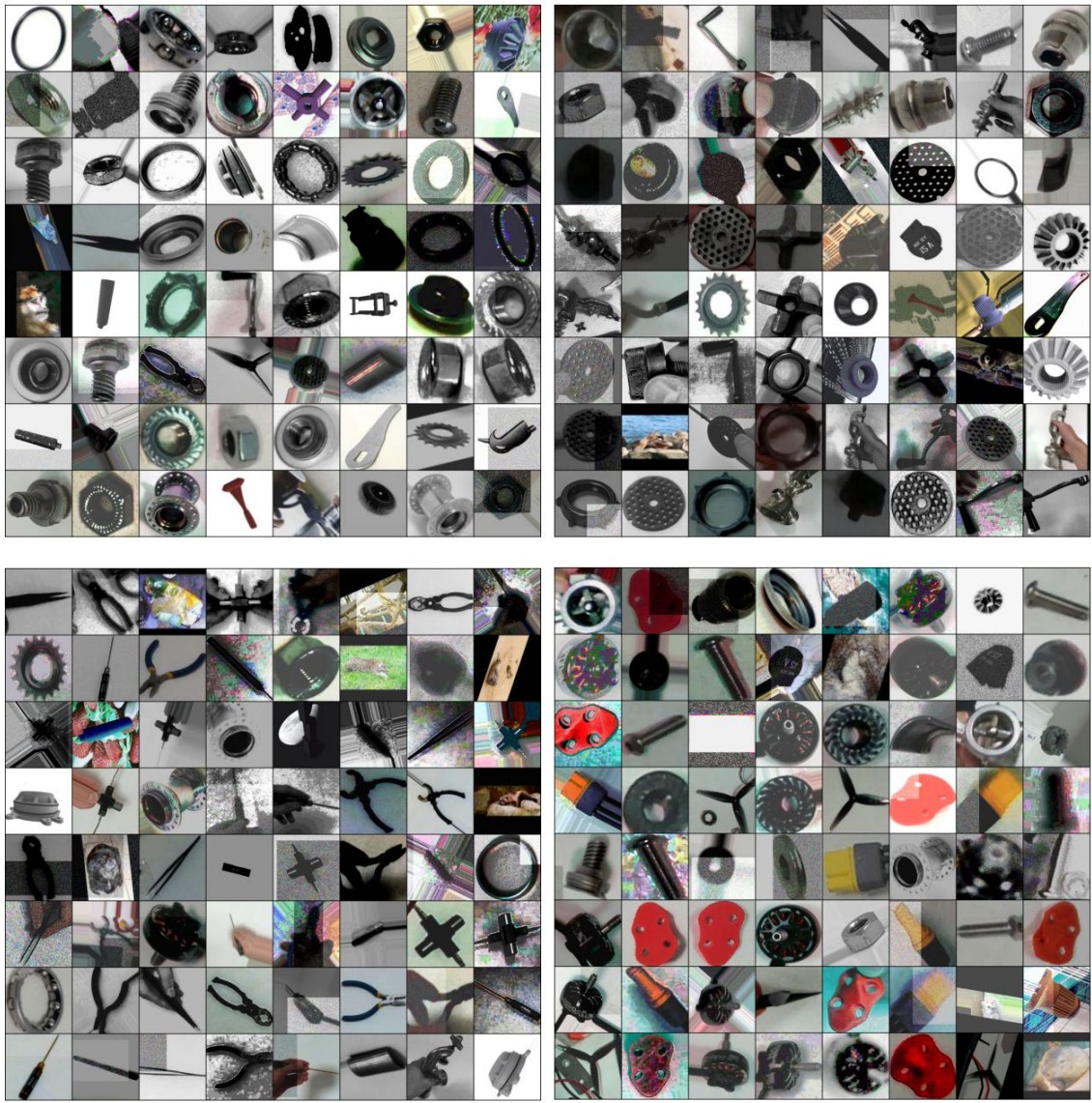

Рис. 3. Примеры аугментированных изображений

## 6.2. Обучение модели

Для всех датасетов мы обучали модели 20 эпох с применением ранней остановки. Для применения метода hard examples mining мы постепенно изменяли долю наиболее сложных примеров, к которым применяется дополнительный штраф, используя линейное расписание: доля уменьшалась линейно от 1.0 на 2 эпохе до 0.1 на 7 эпохе и далее.

Для балансировки классов в тренировочном датасете мы использовали взвешенное семплирование. Так как мы вводим дополнительный класс для посторонних изображений, мы экспериментально подобрали долю этого класса равную 1/3, оставшиеся 2/3 были поровну разделены между всеми остальными классами.

В качестве оптимизатора мы применяли AdamW с параметрами weight_decay=0.01, lr=0.005.

## 7. Результаты

## 7.1. Метрики

| Модель | Balanced Accuracy | F1 | Precision | Recall | mAP | Precision@k | mAP@k |
|---|---|---|---|---|---|---|---|
| Bicycle Parts | | | | | | | |
| Swin Transformer | 0.9858 | 0.9777 | 0.9704 | 0.9858 | 0.9971 | 0.9814 | 0.9822 |
| MobileNetV3 | 0.9744 | 0.9761 | 0.9784 | 0.9744 | 0.9959 | 0.9825 | 0.9830 |
| EfficientNetV2 | **0.9919** | **0.9860** | **0.9806** | **0.9919** | **0.9981** | **0.9897** | **0.9903** |
| Grinder Details | | | | | | | |
| Swin Transformer | 0.9745 | 0.9749 | 0.9754 | 0.9745 | 0.9961 | **0.9679** | 0.9681 |
| MobileNetV3 | **0.9776** | **0.9802** | **0.9831** | **0.9776** | 0.9958 | 0.9769 | **0.9771** |
| EfficientNetV2 | 0.9710 | 0.9752 | 0.9796 | 0.9710 | **0.9970** | 0.9740 | 0.9742 |
| Drone Tools | | | | | | | |
| Swin Transformer | 0.9305 | 0.9282 | 0.9306 | 0.9305 | 0.9808 | 0.9426 | 0.9427 |
| MobileNetV3 | 0.9147 | 0.9228 | 0.9378 | 0.9147 | 0.9720 | 0.9478 | 0.9481 |
| EfficientNetV2 | **0.9414** | **0.9486** | **0.9584** | **0.9414** | **0.9899** | **0.9584** | **0.9582** |
| Drone Details | | | | | | | |
| Swin Transformer | **0.9877** | **0.9894** | **0.9911** | **0.9877** | **0.9992** | 0.9841 | 0.9842 |
| MobileNetV3 | 0.9661 | 0.9766 | 0.9887 | 0.9661 | 0.9967 | **0.9901** | **0.9902** |
| EfficientNetV2 | 0.9774 | 0.9824 | 0.9882 | 0.9774 | 0.9987 | 0.9819 | 0.9821 |
| All Details | | | | | | | |
| Swin Transformer | **0.9919** | **0.9909** | **0.9904** | **0.9919** | **0.9978** | **0.9937** | **0.9938** |
| MobileNetV3 | 0.9816 | 0.9823 | 0.9834 | 0.9816 | 0.9932 | 0.9863 | 0.9863 |
| EfficientNetV2 | 0.9717 | 0.9715 | 0.9726 | 0.9717 | 0.9945 | 0.9704 | 0.9718 |

Таблица 2. Метрики обученных моделей на тестовых датасетах

Как можно видеть из таблицы, на тестовых подмножествах наших датасетов модели показывают близкие к идеалу результаты по рассмотренным метрикам. В процессе анализа полученных результатов мы выяснили, что наши наборы данных, несмотря на применение аугментаций, содержат, в основном, не сложные примеры. Далее будут приведены примеры работы моделей на сборочном столе в условиях, моделирующих реальные, а также значения метрик.

## 7.2. Производительность

В таблице 3 для всех использованных нами предобученных моделей компьютерного зрения приведены число параметров модели и GFLOPs ($10^6$ **FL**oating **OP**eration**s**), требуемые для обработки одного изображения размера $3 \times 224 \times 224$.

| Итоговая модель | Params | GFLOPs |
|---|---|---|
| Swin Transformer | 27.526M | 2.321 |
| MobileNetV3 | 2.980M | 0.169 |
| EfficientNetV2 | 20.188M | 2.125 |

Таблица 3. Число параметров и GFLOPs моделей

Также мы провели оценку производительности всей системы, взяв для обработки видео обученную нами модель EfficientNetV2. Мы использовали видеокарту NVIDIA GeForce RTX 3060 Mobile (3840 cores, 120 TMUs, 48 ROPs, 6 GB GDDR6 memory, 192-bit bus width), и однопоточное приложение для обработки видео с камеры с помощью OpenCV. Для обнаружения объектов мы использовали нейронную сеть YOLOv5, дообученную на наших данных для решения этой задачи. Все обнаруженные в кадре объекты обрабатываются моделью EfficientNetV2 одним батчем.

Мы провели все замеры 10 раз, подсчитали выборочные средние и стандартные отклонения, замеры приведены в таблице 4. Здесь "All time" означает время обработки одного кадра всей системой, "Model time" – время обработки батча изображений моделью EfficientNetV2.

| Batch size | All time (ms) mean | All time (ms) std | Model time (ms) mean | Model time (ms) std |
|---|---|---|---|---|
| (65, 3, 192, 192) | 256,2 | 34,78 | 16,6 | 2,07 |
| (44, 3, 192, 192) | 216,6 | 12,65 | 14,6 | 1,07 |
| (32, 3, 192, 192) | 158,1 | 9,56 | 14,9 | 1,20 |
| (16, 3, 192, 192) | 138,8 | 6,34 | 12,7 | 3,06 |
| (14, 3, 192, 192) | 160,2 | 13,46 | 12,9 | 2,02 |
| (5, 3, 192, 192) | 79,6 | 7,18 | 13,9 | 2,47 |
| (2, 3, 192, 192) | 85,1 | 5,97 | 15,6 | 3,84 |

Таблица 4. Время обработки кадра

## 7.3. Построение статистических оценок

Для анализа векторных представлений изображений, полученных с помощью метода Deep Metric Learning, мы построили t-SNE [4] преобразование векторов, а также диаграммы box-and-whisker для выборочных распределений внутриклассовых и межклассовых расстояний, а также оценили вероятности ошибок 1 и 2 рода.

### 7.3.1. Внутриклассовые расстояния

На box-and-whisker диаграммах "Inter-class distances" изображены выборочные распределения внутриклассовых расстояний. Границы ящика соответствуют первому и третьему квантилю, границы усов – квантилям 0.025 и 0.975. Оранжевая линия показывает медиану. Используемое во всех графиках расстояние – евклидова метрика.

### 7.3.2. Межклассовые расстояния

На box-and-whisker диаграммах "Intra-class distances" изображены выборочные распределения межклассовых расстояний (для каждого класса – до любого другого). Использовались те же параметры графиков, что и для "Inter-class distances".

### 7.3.3. Вероятности ошибок 1 и 2 рода.

Для некоторого класса $C_i$ рассмотрим классификатор на основе попарных расстояний с нулевой гипотезой $H_0$, относящий к классу $C_i$ объекты $x$, имеющие среднее расстояние до объектов выбранного класса меньше квантиля $q$:

$$\frac{1}{|\{y: class(y) = C_i\}|} \sum_{y:class(y)= C_i} ||x - y|| < q, \quad H_0: \{class(x) = C_i\}$$

Рассмотрев $q = q_{1-\alpha}$ – квантиль порядка $1 - \alpha$ распределения среднего расстояния от объекта класса $C_i$ до других объектов класса $C_i$, мы получим классификатор с вероятностью ошибки первого рода $\approx \alpha$, а оценку вероятности ошибки второго рода можно получить, посчитав вероятностную массу хвоста распределения.

Рассмотрев $q = q_\beta$ – квантиль порядка $\beta$ распределения среднего расстояния от объекта класса $C_i$ до объектов других классов $C_j, i \neq j$, мы получим классификатор с вероятностью ошибки второго рода ≈ $\beta$, а оценку вероятности ошибки первого рода можно получить, посчитав вероятностную массу хвоста распределения.

Мы выбрали $\alpha = \beta = 0.025$.

### 7.4. Результаты на датасете "Bicycle Parts"

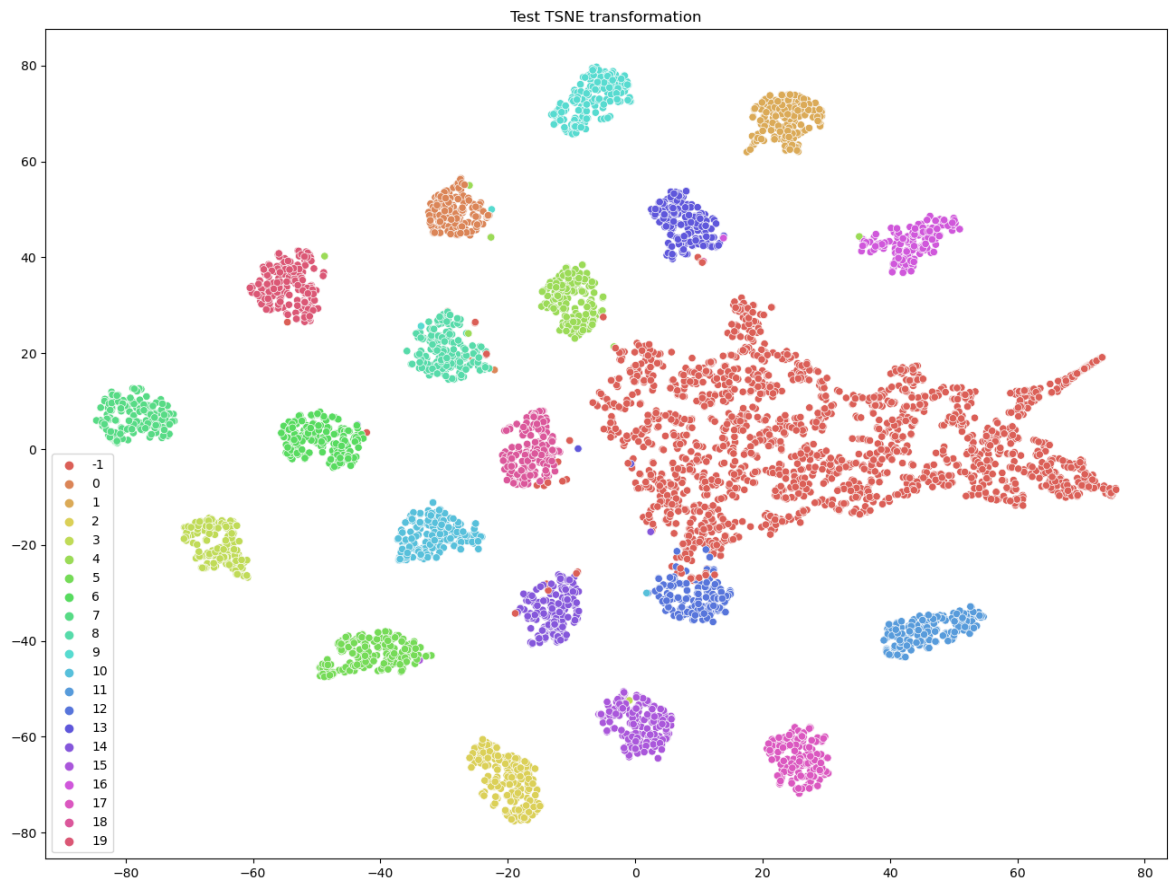

Рис. 4. t-SNE преобразование векторных представлений изображений для датасета "Bicycle Parts", полученных с помощью модели EfficientNetV2

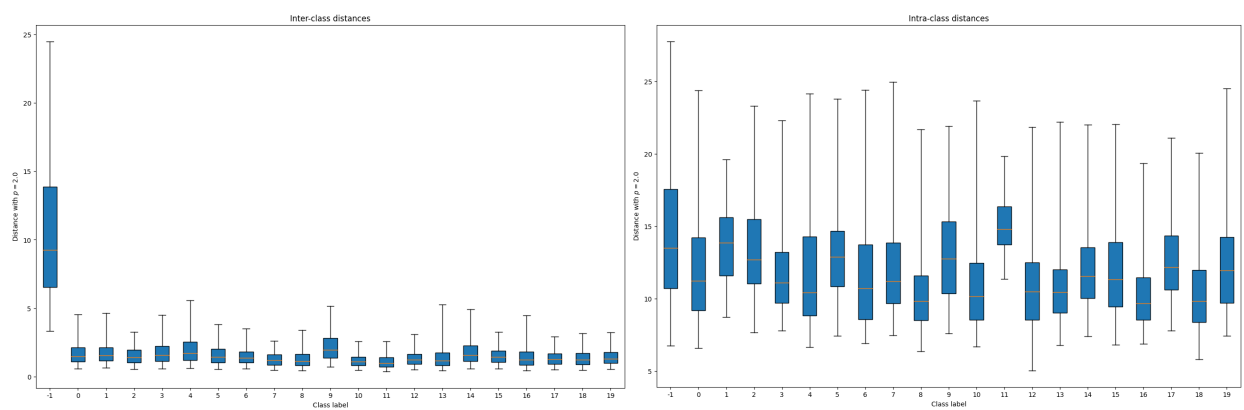

Рис. 5. Box-and-whiskers диаграммы для внутриклассовых и межклассовых расстояний, датасет "Bicycle Parts"

| Класс | По $q_{1-\alpha}$ (внутриклассовые расстояния) | | По $q_\beta$ (межклассовые расстояния) | |
|---|---|---|---|---|
| | 1 рода | 2 рода | 1 рода | 2 рода |
| 0  | 0.025 | 0.006 | 0.002 | 0.025 |
| 1  |       | 0.001 | 0.000 |       |
| 2  |       | 0.000 | 0.000 |       |
| 3  |       | 0.004 | 0.000 |       |
| 4  |       | 0.012 | 0.009 |       |
| 5  |       | 0.000 | 0.001 |       |
| 6  |       | 0.003 | 0.000 |       |
| 7  |       | 0.000 | 0.000 |       |
| 8  |       | 0.002 | 0.000 |       |
| 9  |       | 0.003 | 0.003 |       |
| 10 |       | 0.000 | 0.000 |       |
| 11 |       | 0.000 | 0.000 |       |
| 12 |       | 0.000 | 0.010 |       |
| 13 |       | 0.010 | 0.003 |       |
| 14 |       | 0.002 | 0.001 |       |
| 15 |       | 0.003 | 0.000 |       |
| 16 |       | 0.007 | 0.002 |       |
| 17 |       | 0.000 | 0.000 |       |
| 18 |       | 0.002 | 0.001 |       |
| 19 |       | 0.000 | 0.000 |       |

Таблица 5. Оценки вероятностей ошибок 1 и 2 рода, датасет "Bicycle Parts"

## 7.5. Результаты на датасете "Grinder Details"

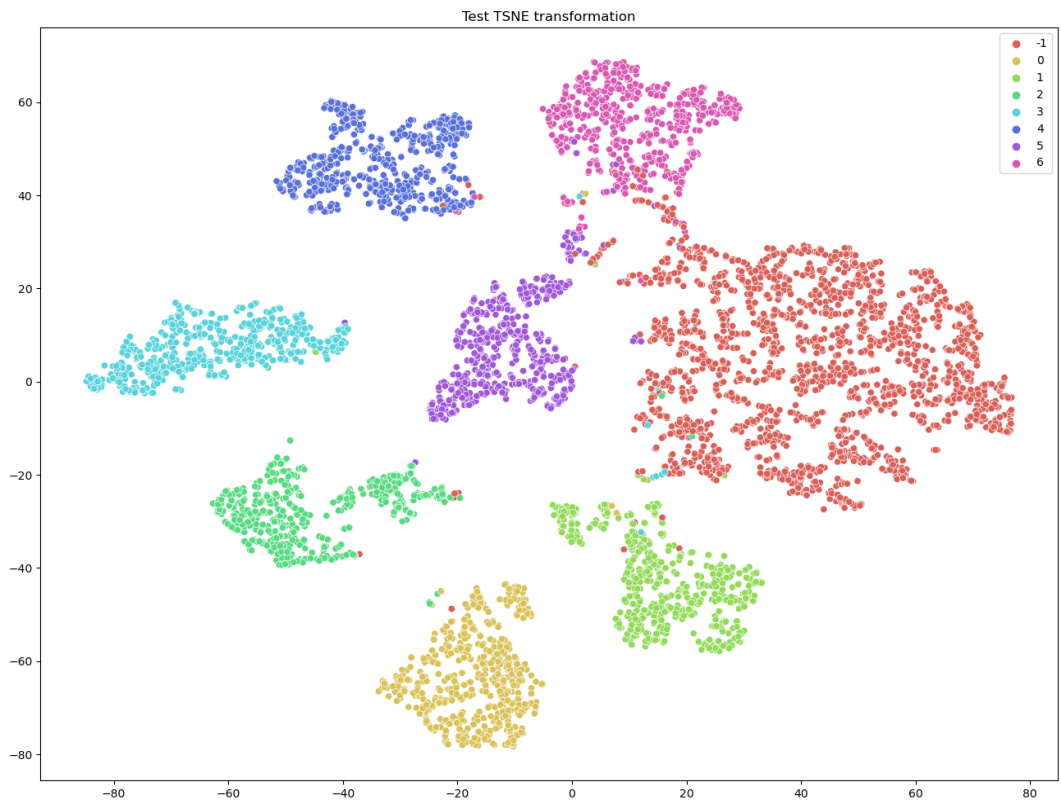

Рис. 6. t-SNE преобразование векторных представлений изображений для датасета "Grinder Details", полученных с помощью модели MobileNetV3

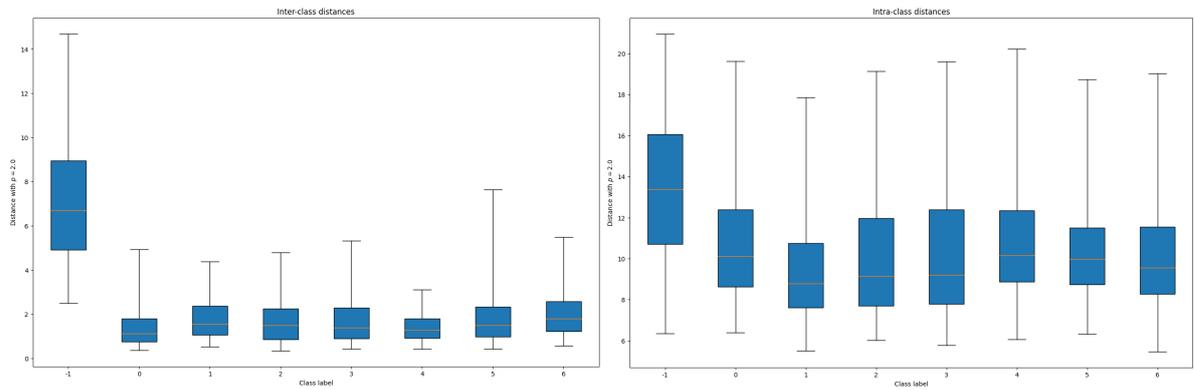

Рис. 7. Box-and-whiskers диаграммы для внутриклассовых и межклассовых расстояний, датасет "Grinder Details"

| Класс | По $q_{1-\alpha}$(внутриклассовые расстояния) | | По $q_\beta$(межклассовые расстояния) | |
|---|---|---|---|---|
| | 1 рода | 2 рода | 1 рода | 2 рода |
| 0 | 0.025 | 0.008 | 0.005 | 0.025 |
| 1 | | 0.007 | 0.006 | |
| 2 | | 0.009 | 0.004 | |
| 3 | | 0.018 | 0.015 | |
| 4 | | 0.000 | 0.000 | |
| 5 | | 0.051 | 0.103 | |
| 6 | | 0.025 | 0.025 | |

Таблица 6. Оценки вероятностей ошибок 1 и 2 рода, датасет "Grinder Details"

## 7.6. Результаты на датасете "Drone Tools"

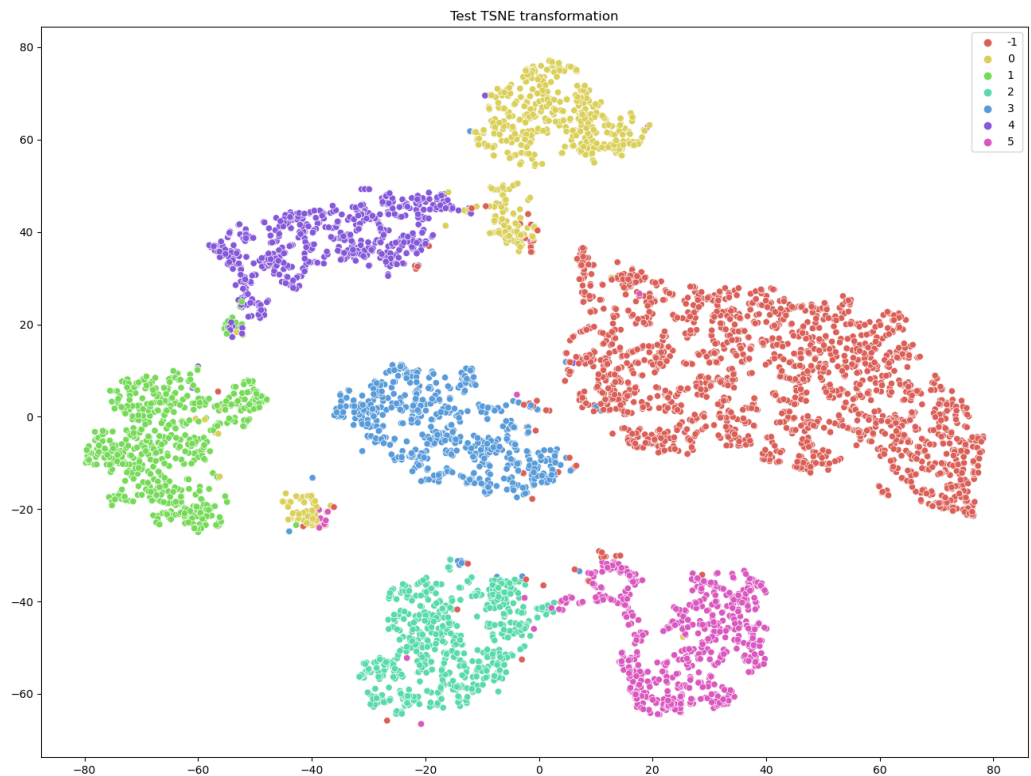

Рис. 8. t-SNE преобразование векторных представлений изображений для датасета "Drone Tools", полученных с помощью модели EfficientNetV2

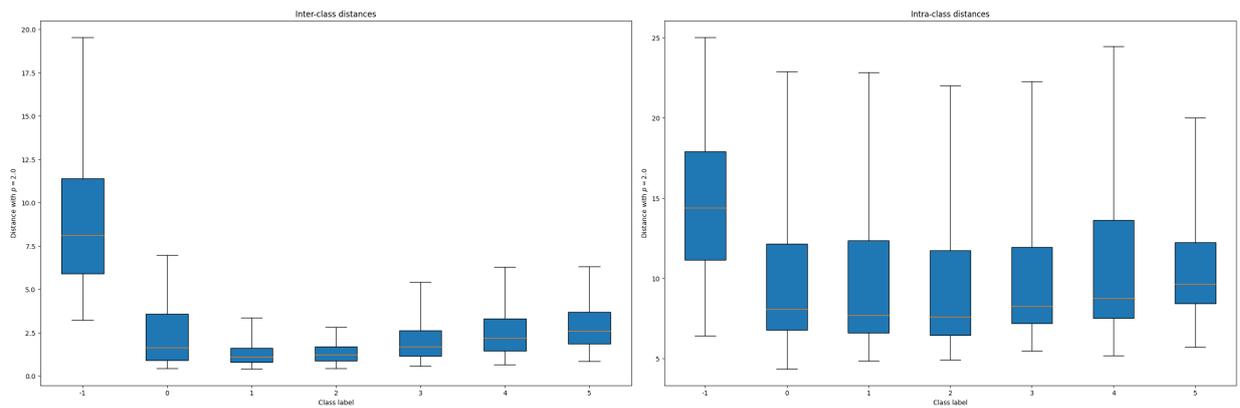

Рис. 9. Box-and-whiskers диаграммы для внутриклассовых и межклассовых расстояний, датасет "Drone Tools"

| Класс | По $q_{1-\alpha}$(внутриклассовые расстояния) | | По $q_\beta$(межклассовые расстояния) | |
|---|---|---|---|---|
| | 1 рода | 2 рода | 1 рода | 2 рода |
| 0 | 0.025 | 0.209 | 0.292 | 0.025 |
| 1 | | 0.005 | 0.010 | |
| 2 | | 0.000 | 0.002 | |
| 3 | | 0.024 | 0.022 | |
| 4 | | 0.068 | 0.066 | |
| 5 | | 0.050 | 0.042 | |

Таблица 7. Оценки вероятностей ошибок 1 и 2 рода, датасет "Drone Tools"

## 7.7. Результаты на датасете "Drone Details"

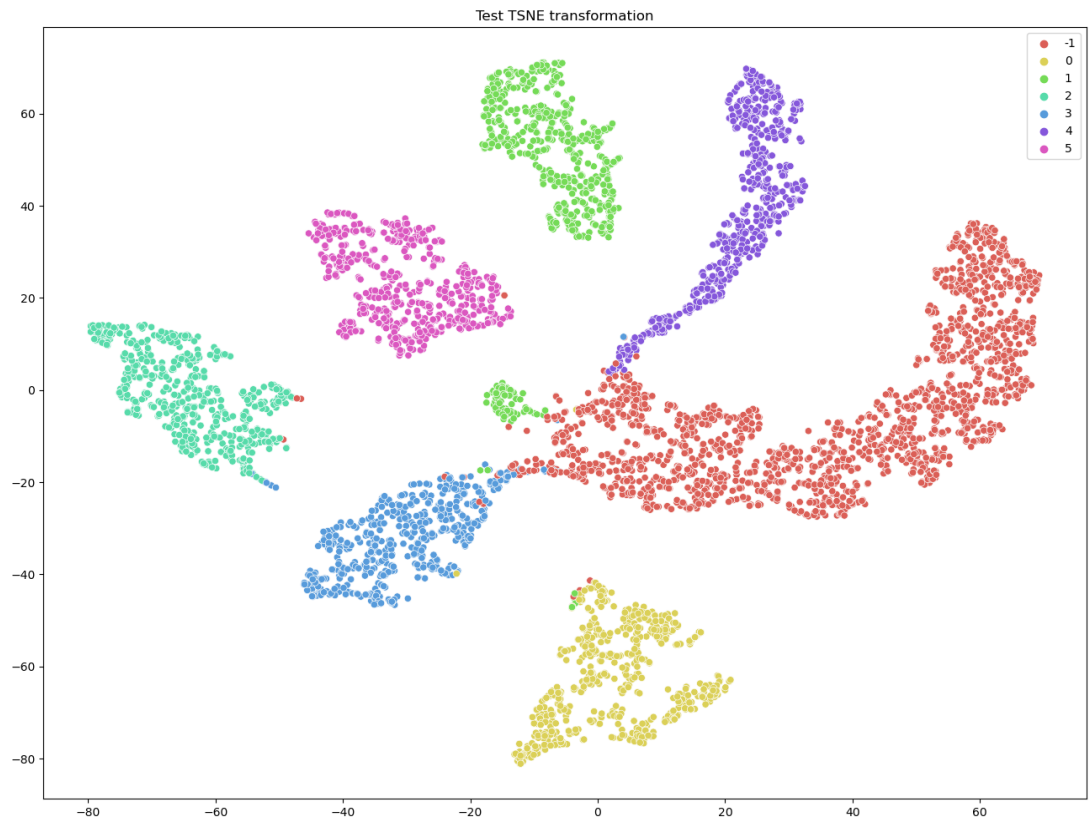

Рис. 10. t-SNE преобразование векторных представлений изображений для датасета "Drone Details", полученных с помощью модели Swin Transformer

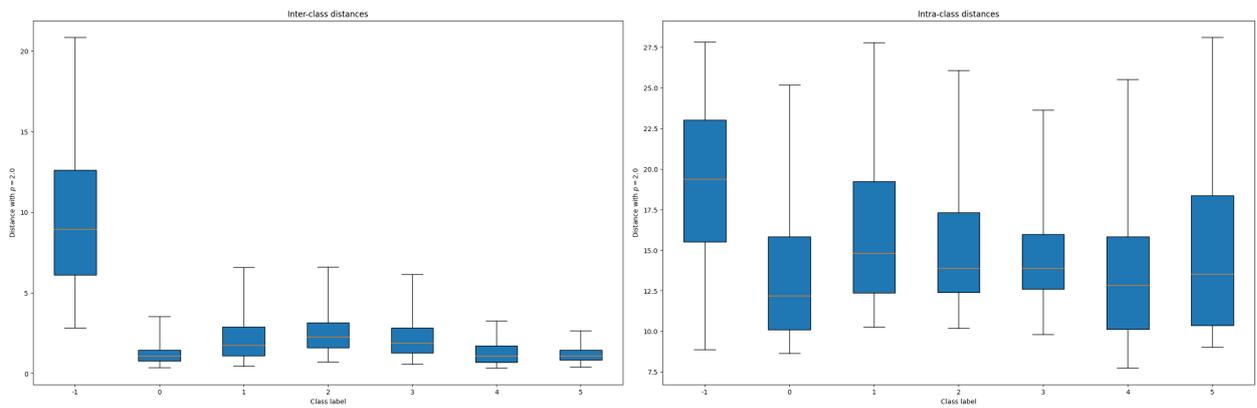

Рис. 11. Box-and-whiskers диаграммы для внутриклассовых и межклассовых расстояний, датасет "Drone Details"

| Класс | По $q_{1-\alpha}$(внутриклассовые расстояния) | | По $q_\beta$(межклассовые расстояния) | |
|---|---|---|---|---|
| | 1 рода | 2 рода | 1 рода | 2 рода |
| 0 | 0.025 | 0.001 | 0.000 | 0.025 |
| 1 | | 0.002 | 0.001 | |
| 2 | | 0.002 | 0.003 | |
| 3 | | 0.003 | 0.002 | |
| 4 | | 0.001 | 0.003 | |
| 5 | | 0.000 | 0.000 | |

Таблица 8. Оценки вероятностей ошибок 1 и 2 рода, датасет "Drone Details"

## 7.8. Результаты на датасете "All Details"

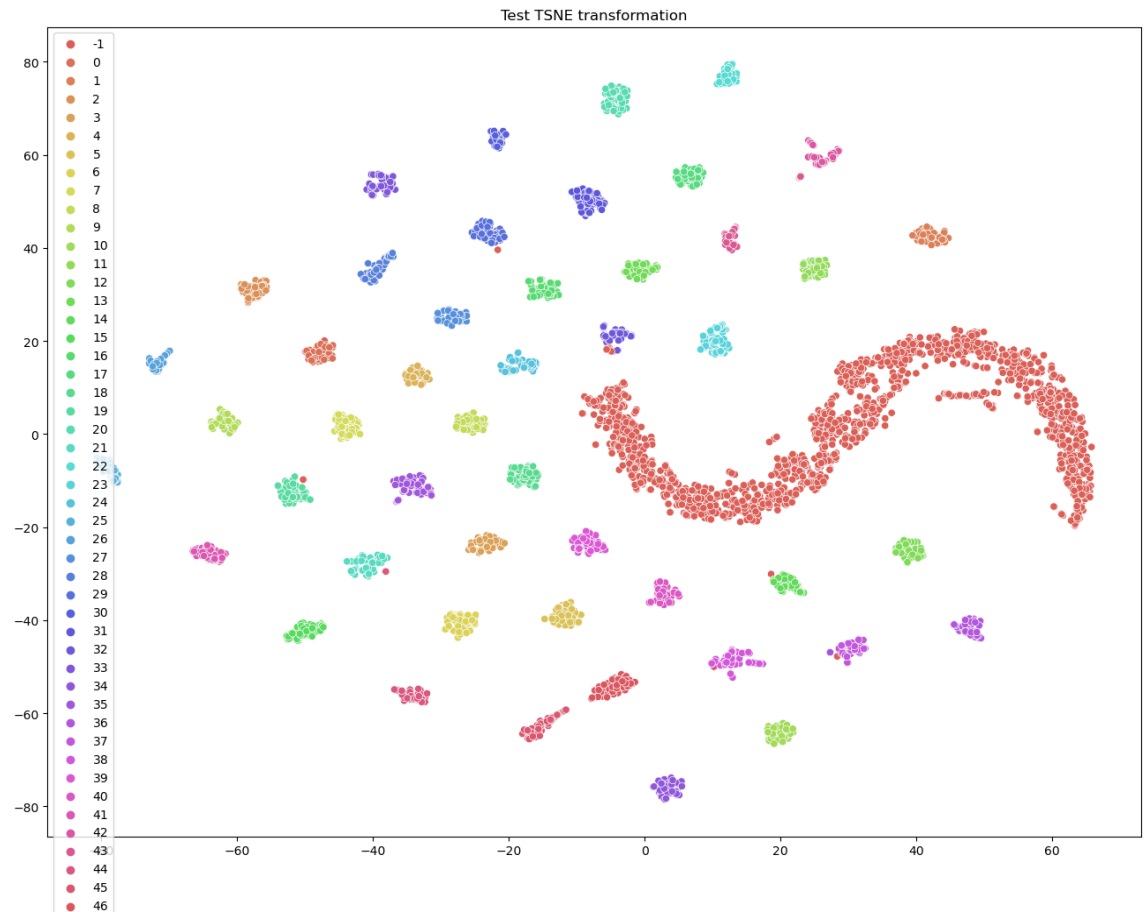

Рис. 12. t-SNE преобразование векторных представлений изображений для датасета "All Details", полученных с помощью модели Swin Transformer

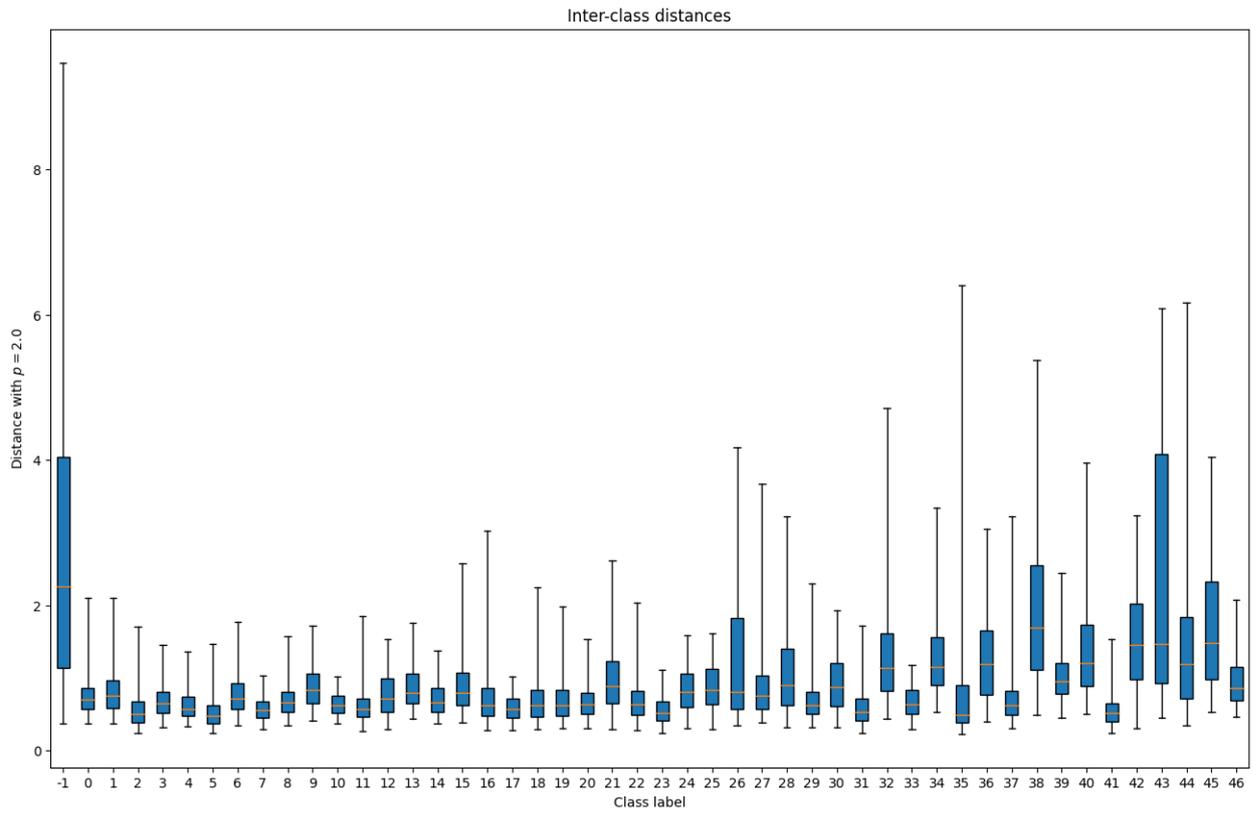

Рис. 13. Box-and-whiskers диаграмма для внутриклассовых расстояний, датасет "All Details"

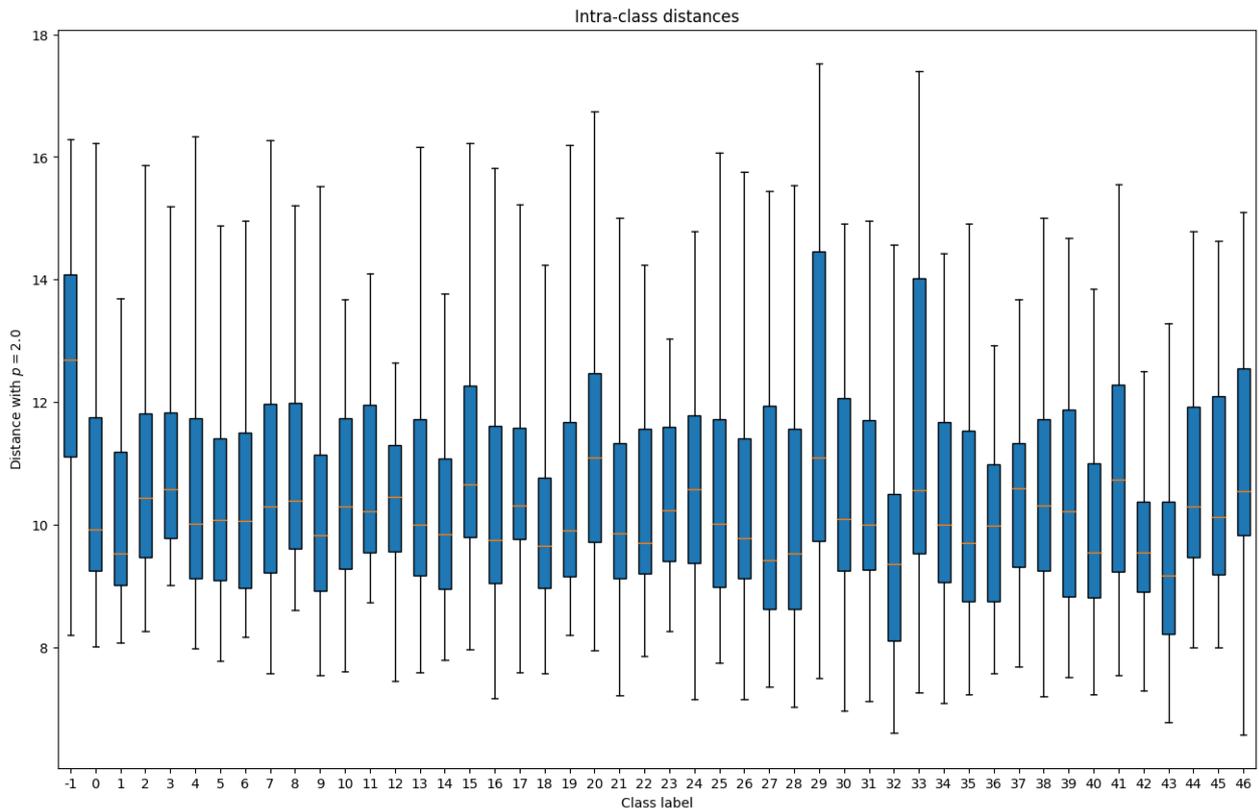

Рис 14. Box-and-whiskers диаграмма для межклассовых расстояний, датасет "All Details"

| Класс | По $q_{1-\alpha}$ | | По $q_\beta$ | | Класс | По $q_{1-\alpha}$ | | По $q_\beta$ | |
|---|---|---|---|---|---|---|---|---|---|
| | 1 рода | 2 рода | 1 рода | 2 рода | | 1 рода | 2 рода | 1 рода | 2 рода |
| 0 | | 0.000 | 0.000 | | 24 | | 0.000 | 0.000 | |
| 1 | | 0.000 | 0.000 | | 25 | | 0.000 | 0.000 | |
| 2 | | 0.000 | 0.000 | | 26 | | 0.000 | 0.000 | |
| 3 | | 0.000 | 0.000 | | 27 | | 0.000 | 0.000 | |
| 4 | | 0.000 | 0.000 | | 28 | | 0.000 | 0.000 | |
| 5 | | 0.000 | 0.000 | | 29 | | 0.000 | 0.000 | |
| 6 | | 0.000 | 0.000 | | 30 | | 0.000 | 0.000 | |
| 7 | | 0.000 | 0.000 | | 31 | | 0.000 | 0.000 | |
| 8 | | 0.000 | 0.000 | | 32 | | 0.000 | 0.000 | |
| 9 | | 0.000 | 0.000 | | 33 | | 0.000 | 0.000 | |
| 10 | | 0.000 | 0.000 | | 34 | | 0.000 | 0.000 | |
| 11 | 0.025 | 0.000 | 0.000 | 0.025 | 35 | 0.025 | 0.000 | 0.000 | 0.025 |
| 12 | | 0.000 | 0.000 | | 36 | | 0.000 | 0.000 | |
| 13 | | 0.000 | 0.000 | | 37 | | 0.000 | 0.000 | |
| 14 | | 0.000 | 0.000 | | 38 | | 0.020 | 0.001 | |
| 15 | | 0.000 | 0.000 | | 39 | | 0.000 | 0.000 | |
| 16 | | 0.000 | 0.000 | | 40 | | 0.000 | 0.000 | |
| 17 | | 0.000 | 0.000 | | 41 | | 0.000 | 0.000 | |
| 18 | | 0.000 | 0.000 | | 42 | | 0.000 | 0.000 | |
| 19 | | 0.000 | 0.000 | | 43 | | 0.001 | 0.009 | |
| 20 | | 0.000 | 0.000 | | 44 | | 0.000 | 0.001 | |
| 21 | | 0.000 | 0.000 | | 45 | | 0.000 | 0.002 | |
| 22 | | 0.000 | 0.000 | | 46 | | 0.000 | 0.000 | |
| 23 | | 0.000 | 0.000 | | | | | | |

Таблица 9. Оценки вероятностей ошибок 1 и 2 рода, датасет "All Details"

## 8. Анализ результатов работы моделей

Помимо тестовой части наших датасетов, мы протестировали наши модели на фотографиях, сделанных непосредственно в процессе сборки. В этом разделе мы привели результаты работы моделей на этих фотографиях.

### 8.1. Результаты модели детекции

Для детекции объектов на изображении мы использовали модель YOLOv5, дообученную на собранном нами наборе данных, содержащем изображения различных объектов и разметку с их ограничивающими рамками.

Текущая модель показывает достаточно хорошие результаты в условиях, когда камера расположена перпендикулярно над столом, но качество работы значительно ухудшается при расположении камеры около стола, под углом. В целом, она обнаруживает около половины интересующих нас объектов. В дальнейшем мы планируем собрать больше данных для обучения модели детекции, а также попробовать другие модели, например, YOLOv8.

### 8.2. Результаты модели классификации

Мы сравнивали наши результаты с работами Deep Quadruplet Network [2], QuadNet [17] и FIDI [13], в которых авторы решают задачу re-identification, также используя различные комбинированные функции потерь. Их авторы используют датасеты CUHK, VIPeR,

Market1501, DukeMTMC, созданные для решения задачи person re-identification, а также датасет MVB – для задачи baggage re-identification.

Так как в разных подходах используются различные архитектуры моделей, функции потерь, и метрики, при сравнении своих результатов с этими статьями мы рассматриваем метрику accuracy, значения которой есть во всех статьях для всех датасетов.

Обученные нами модели показывают на наших задачах результаты (в Таблице 10), сравнимые по выбранной метрике accuracy с результатами из работ [2], [17], [13]. Таким образом, мы получили модель классификации, способную обрабатывать посторонние объекты, определяя их в специальный класс, и сохраняющую при этом качество работы для остальных классов.

Стоит отметить, что приведённые в таблице 10 результаты отличаются в худшую сторону от результатов работы моделей на тестовой части датасетов (таблица 2). Это следствие отсутствия сложных примеров в наших датасетах. Для улучшения результатов работы модели в процессе сборки, мы планируем расширить наши наборы данных.

| Модель | Датасет | Accuracy |
|---|---|---|
| Deep Quadruplet Network | CUHK03 | 74.47 % |
|  | CUHK01(p=486) | 62.55% |
|  | CUHK01(p=100) | 79.00% |
|  | VIPeR | 48.42% |
| QuadNet | MVB | 80.30% |
| FIDI | Market1501 | 94.50% |
|  | DukeMTMC | 88.10% |
|  | CUHK03-D | 72.10% |
|  | CUHK03-L | 75.00% |
| EfficientNetV2 (Ours) | Bicycle Parts (Ours) | 70.14% |
| MobileNetV3 (Ours) | Grinder Details (Ours) | 98.13% |
| Swin Transformer (Ours) | All Details (Ours) | 66.81% |

Таблица 10. Сравнение наших моделей с другими работами

8.3. Полученные векторные представления

Чтобы визуально оценить построенные векторные представления изображений, с помощью которых далее решается задача классификации, мы построили графики t-SNE преобразования этих векторов. Для всех наших датасетов большинство объектов группируются в легко отделимые кластеры. Объекты, лежащие на границе своего кластера или являющиеся выбросами из него могут быть ошибочно отнесены к другому классу на основе кластеризации. Чтобы оценить долю таких объектов, а также степень их близости к другим классам, мы построили диаграммы box-and-whiskers, а также посчитали оценки вероятностей ошибок 1 и 2 рода простого классификатора.

Проанализировав диаграммы box-and-whiskers можно заметить, что для некоторых классов построенные векторные представления имеют существенно больший разброс попарных расстояний , чем для всех остальных. В частности, это заметно для класса "-1" (посторонние объекты), это следствие большого разнообразия объектов в классе. Например, для наших наборов данных, как фотография коробки, так и фотография животного будут относиться к этому классу. Для других классов высокий разброс попарных расстояний может указывать на плохое качество полученных векторов изображений данного класса.

С помощью таблиц оценок ошибок 1 и 2 рода можно оценить качество построенных векторных представлений. Несмотря на то, что мы в нашей модели используем более сложный классификатор, чем решающее правило на основе квантиля распределения попарных расстояний, полученные оценки позволяют выявить проблемные для построения векторных представлений классы, для которых может потребоваться собрать больше данных.

9. Благодарности

10. Список литературы